\documentclass[journal]{IEEEtran}
\ifCLASSINFOpdf
\else
\fi
\usepackage{amsmath}
\usepackage{algorithm}
\usepackage{algpseudocode}
\usepackage{graphicx}
\usepackage{xcolor}
\usepackage{multirow}
\usepackage{amsfonts}
\usepackage{amssymb}

\usepackage{array}

\newtheorem{theorem}{\textbf{Theorem}}

\newtheorem{remark}{\textbf{Remark}}

\newtheorem{assumption}{\textbf{Assumption}}

\begin{document}
%
\title{Actor-Critic Reinforcement Learning \\with Phased Actor} 
%

%
%

\author{Ruofan~Wu,
        Junmin~Zhong,
        and~Jennie~Si,~\IEEEmembership{Fellow,~IEEE}
}

\maketitle

\begin{abstract}
Policy gradient methods in actor-critic reinforcement learning (RL) have become perhaps the most promising approaches to solving continuous optimal control problems. However, the  trial-and-error nature of RL and the inherent randomness associated with solution approximations cause variations in the learned optimal values and policies. This has significantly hindered their successful deployment in real life applications  where control responses need to meet dynamic performance criteria deterministically. 
Here we propose a novel phased actor in actor-critic (PAAC) method, aiming at improving policy gradient estimation and thus the quality of the control policy. Specifically, PAAC accounts for both $Q$ value and TD error in its actor update. 
We prove qualitative properties of PAAC for learning convergence of the value and policy, solution  optimality, and stability of system dynamics. Additionally, we show variance reduction in policy gradient estimation. PAAC performance is systematically and quantitatively evaluated in this study  using DeepMind Control Suite (DMC). Results show that PAAC leads to  significant performance improvement measured by total cost, learning variance, robustness, learning speed and success rate. As PAAC can be piggybacked onto general policy gradient learning frameworks, we select  well-known methods such as direct heuristic dynamic programming (dHDP), deep deterministic policy gradient (DDPG) and their variants to demonstrate the effectiveness of PAAC. Consequently we provide a unified view  on these related policy gradient algorithms.
\end{abstract}

\begin{IEEEkeywords}
Actor-critic reinforcement learning, policy gradient, learning variance, direct heuristic dynamic programming (dHDP), deep determinstic policy gradient (DDPG)

\end{IEEEkeywords}

%
\IEEEpeerreviewmaketitle

\section{Introduction}

Many tasks of interest, most notably physical control tasks ranging from robotics, industrial processes to ground and air vehicles, have continuous (real valued) state and action spaces. Actor-critic policy gradient reinforcement learning methods have become the most feasible solution framework to address this class of  challenging yet highly relevant problems in real life. Recently, some great progress has been made in deep reinforcement learning (DRL) addressing control problems involving high-dimensional, continuous states and control spaces. Several algorithms, such as deep deterministic policy gradient (DDPG) \cite{lillicrap2015continuous}, soft actor critic (SAC) \cite{haarnoja2018soft}, D4PG \cite{haarnoja2018soft} and twin delayed DDPG (TD3) \cite{fujimoto2018addressing}, to name some, have demonstrated their potentials. However, there still are major obstacles for DRL methods to be effectively implemented and deployed in real time applications. Major technical obstacles  include the high variance problem \cite{bhatnagar2009natural,bjorck2021high}, low data efficiency \cite{dulac2021challenges}, slow learning \cite{johnson2013accelerating}, or even instability of learning, and poor task performance \cite{henderson2018deep,duan2016benchmarking}. Thus, innovative solutions are critically needed to address these challenges in RL control designs.

To improve  DRL learning performance, the majority of existing approaches focuses on improving the quality of critic estimation  within an actor-critic solution framework. Several ideas have been investigated and shown promising. The deterministic policy gradient (DPG) \cite{silver2014deterministic} and deep $Q$ network (DQN) \cite{mnih2013playing,mnih2015human} use a target network, which is a copy of the estimated value function being held up for a number of steps to serve as a stable target in order to reduce the accumulation of approximation error \cite{fujimoto2018addressing}. The n-step methods such as TD($\lambda$) \cite{sutton2018reinforcement} and Q($\sigma$) \cite{de2018multi} represent  another class of approaches that are known to reduce learning  variances. Specifically, rollout methods \cite{bertsekas2010rollout,tesauro1994td,silver2018general} have enabled the success of some of the most important milestone case studies in discrete state and control problems \cite{tesauro1994td,silver2016mastering}. For continuous state and control problems, Rainbow \cite{hessel2018rainbow}, LNSS \cite{zhong2024long} and D4PG \cite{barth2018distributed} have shown n-step learning help speed-up in learning and an improvement of the final performance. The TD3 \cite{fujimoto2018addressing} is based on an idea of using a value estimate with lower variance. It is implemented by double Q learning and shown to result in higher quality policy updates  \cite{hasselt2010double}.

While the above methods are promising in terms of improving critic estimation performance, those methods aiming at improving actor network training rely on using advantage functions, namely, the difference between the $Q$ value and the state value. These result are mostly developed under the MDP framework and for stochastic policies. 
Advantage is considered a baseline method that helps reduce variance for gradient-based policy searches \cite{arulkumaran2017deep}. Greensmith, et al. \cite{greensmith2004variance} provided a proof of using a baseline in actor-critic for policy gradient estimation toward variance reduction. Studies also have shown that a baseline that is independent of the action will affect the variance without changing the expected value of the policy gradient \cite{sutton1999policy,greensmith2004variance}.
Expanded from advantage ideas, generalized advantage estimators (GAEs) proposed in A2C \cite{mnih2016asynchronous,schulman2015high} uses a discounted sum of Bellman residuals to tradeoff between the bias and the variance by introducing a hyperparameter $\lambda$ similar to that in TD($\lambda$).
Policy gradients with parameter-based exploration (PGPE) \cite{sehnke2010parameter} applies a moving-average baseline to reduce variance. 
An improved PGPE \cite{zhao2011analysis} is developed to  optimize the baseline by minimizing  the variance in policy gradient estimation. A more recent study \cite{wuvariance} proposes an action-dependent factorized baseline to further reduce the variance.
In the above methods, since the true state value is not known, TD error has been used to approximate the advantage function \cite{baird1993advantage,kakade2001natural,degris2012off}. Additionally, it has been shown that one step TD error is an unbiased estimate of the advantage function \cite{bhatnagar2009natural}, and it has been used in other actor-critic methods \cite{schulman2015high,peters2008natural,peters2003reinforcement}. Evaluations have shown the effectiveness of using TD in  actor learning to  dramatically reduce the variance of the gradient estimates \cite{bhatnagar2007incremental}.

As discussed in the above, advantage function or TD error has been effectively utilized in stochastic actor-critic methods, especially in the context of finite MDPs. 
For actor-critic algorithms that address deterministic continuous problems, a common approach to policy gradient estimation centers on using $Q$ value, a method that directly leads to policy gradient algorithms such as  DDPG \cite{lillicrap2015continuous}, TD3 \cite{fujimoto2018addressing} and DQN \cite{mnih2013playing}. While $Q$ value is straight-forward to use in policy gradient, it may introduce higher variance in policy gradient \cite{schulman2015high,baird1993advantage}. Even though studies have shown 
great variance reduction using TD error \cite{sutton2018reinforcement,williams1992simple,kimura1998analysis} as it is a consistent estimate of the advantage function, TD error only measures the discrepancy between the predicted value and the target value, which may not be able to efficiently guide exploration, which is much needed in reaching optimal policy. 
Insufficient exploration by using TD error alone is observed in DPG given its deterministic nature \cite{silver2014deterministic}.

To bridge the important knowledge gap between existing policy gradient estimation methods and what is critically needed for deterministic policy gradient algorithms in many controls problems, we introduce a new phased actor in actor-critic (PAAC) method for improved learning performance facilitated through the actor network. PAAC aims at reducing policy gradient variance while encouraging exploration, and consequently leading to improved learning performance. The phased switching mechanism in PAAC allows for the policy gradient estimator to transition from favoring $Q$ value to TD error as learning proceeds over time. We expect reduced variance in gradient estimation  by using the TD error at later stage of learning and using the $Q$ value during the initial phase of learning to encourage sufficient exploration in the policy space. 

It is important to note that the new PAAC method can be implemented and integrated into general policy gradient learning algorithms. We select well-known algorithms such as direct heuristic dynamic programming (dHDP), deep deterministic policy gradient (DDPG), and a series of their variants in evaluations to demonstrate significant performance improvement by using PAAC. On one hand, these methods are comparable in their basic structures and thus the comparisons are meaningful. On the other hand, these methods are closely related to more complex methods such as TD3 and D4PG, which have more bells and whistles built in, the effect of which may confound that of integrating PAAC.

The contributions of this study include the following. 
1) The new PAAC method addresses the important deterministic actor-critic learning problem. It can be piggybacked onto general policy gradient methods under an actor-critic RL framework to enable stronger performances of leading algorithms. 
2) We provide analytical results on learning performance guarantees such as learning convergence of value and policy, solution optimality, and stability of system dynamics. We also show variance reduction in the policy gradient estimation. 
3) Our systematic evaluations demonstrate the effectiveness of PAAC on learning performance measured by total cost, learning variance, robustness, learning speed, and success rate using benchmark  tasks in DMC Suit. 
4) Finally, as a byproduct of this study, we have obtained an enhanced version of the basic dHDP actor-critic learning.

\section{Related Work}
\label{sec:related work}

\subsection{A Unified View of Policy Gradient Actor-Critic}
\begin{figure}
    \centering
    \includegraphics[width=250pt]{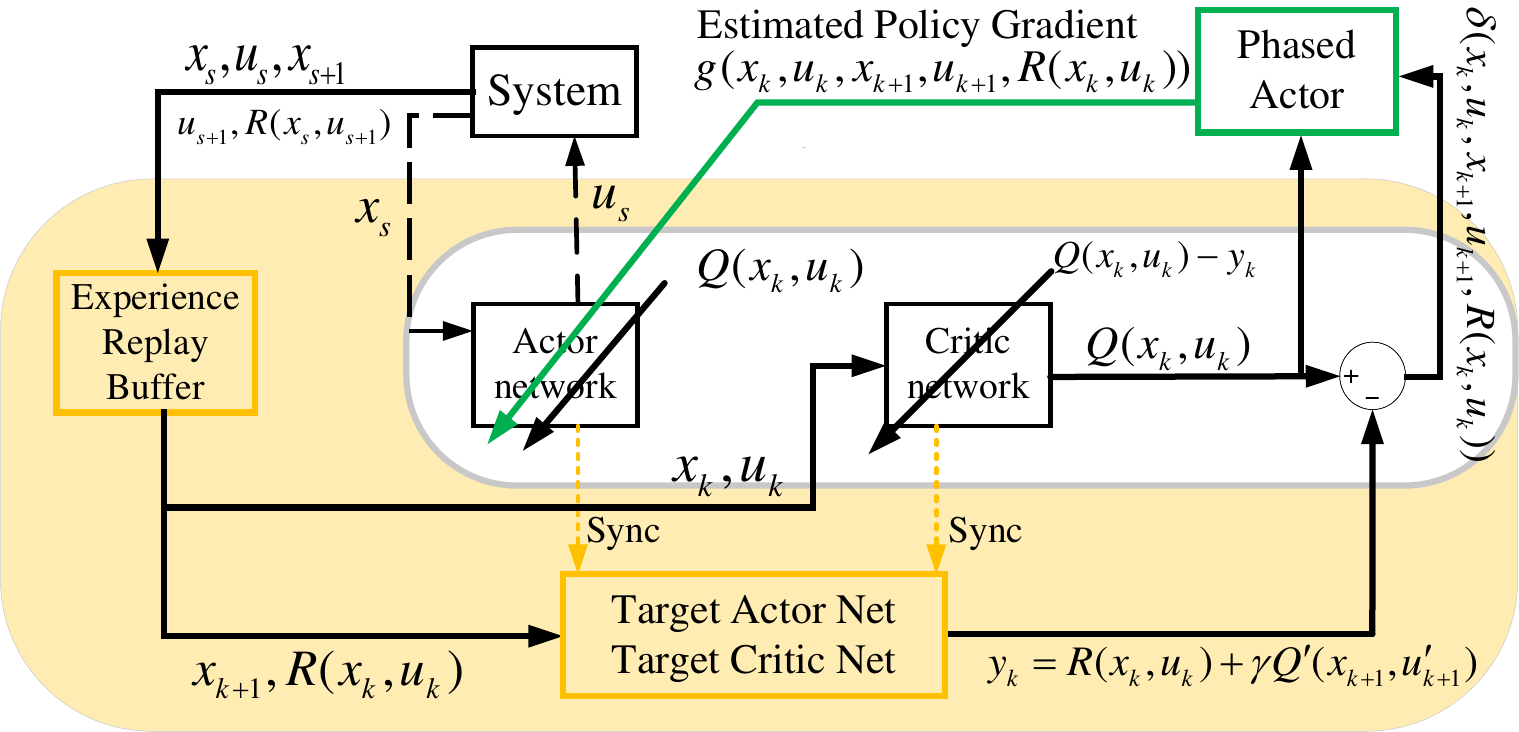}
    \caption{A unified actor-critic framework: Illustration of different DRL algorithms and how they relate to one another.     Inside the white bubble is the vanilla dHDP as in \cite{si2001online}, the yellow bubble forms DDPG \cite{lillicrap2015continuous}.  The green box is how the phased actor is realized, and it can be piggybacked onto general actor-critic structures such as vanilla dHDP and DDPG as shown. The dash lines show the sample collection path and black lines show how  learning proceeds.
}
    \label{fig:algorithm comparison}
\end{figure}

Figure \ref{fig:algorithm comparison} shows a block diagram of dHDP, DDPG and PAAC under a unified policy gradient actor-critic framework. The interaction between the environment and the actor are illustrated in black dashed lines. The critic network provides an estimated $Q$ value during each iteration given state $x_k$ and action $u_k$. The center white box encloses one of the original policy gradient actor-critic structures, which is referred to as vanilla dHDP in this study. Inside the vanilla dHDP structure, the actor is updated by the approximate gradient of $Q$ from the critic network over the actor network parameters while the critic is updated by the TD error. In this study, we include  two additional procedures (shown inside the yellow box) that have been established in contemporary RL for good learning performance: experience replay to improve data efficiency and target networks to provide a more stable Bellman recursion target.
The experience replay buffer containing \{($x_s,u_s$)\} provides sampled pair ($x_k,u_k$) for actor and critic updates. For the target networks, $Q'$ and $u'$ represent target $Q$ and target action values, respectively. The two target networks' parameters are updated based on the respective critic and actor networks in a designated manner. Lastly in Figure \ref{fig:algorithm comparison}, the proposed novel phased actor provides an estimated policy gradient based on  
both $Q$ value and TD error $\delta(x_k,u_k,x_{k+1},R(x_k,u_k))$ in actor updates. 
Detailed discussions on the phased actor are given in section \ref{sec:PAAC} below.

\subsection{dHDP}
\label{sec:related dHDP}
The white box in Figure \ref{fig:algorithm comparison} shows the structure of dHDP 
\cite{si2001online,enns2002apache,enns2003helicopter}. 
It was introduced as an online reinforcement learning algorithm, but with little trouble, it can also be implemented offline or using batch samples \cite{hafner2011reinforcement}. The critic network of dHDP approximates the state-action value function to evaluate the performance of a policy represented by the actor network. In the context of the unified framework, setting the experience replay buffer to 1 corresponds to online dHDP. When used in batch mode, it can be implemented in the same way as using mini batches. 
The critic network of dHDP is updated to reduce the Bellman error according to the Bellman equation (\ref{equ:dt hjb}). The loss function is defined as
\begin{equation}
L_{dHDP}=\left({Q}\left(x_k, u_k\right) - \gamma {Q}(x_{k+1},\pi(x_{k+1}))-R(x_k,u_k)\right)^2,
\end{equation}
where ${Q}\left(x_k, u_k\right)$ is the critic network output, $R(x_k,u_k)$ is the stage cost, and $\pi$ is control policy. Note that for each data tuple $(x_k,u_k,x_{k+1})$, dHDP has the flexibility to update the actor and critic networks for a few steps (can be 1 step per sample pair), respectively before sample another state-action pair for the next learning iteration. 

The dHDP adapts the action network by using the 
error between the desired ultimate cost-to-go and the approximate ${Q}$ value from the critic network. 
The policy gradient estimator for dHDP is expressed as
\begin{equation}
g_{_{dHDP}}=\nabla_\pi {Q}\left(x_k, \pi(x_k)\right).
\end{equation}

\subsection{DDPG} 
\label{sec:ddpg}
The DDPG method  is a model‐free actor‐critic algorithm that has been shown capable of learning different continuous control tasks directly from raw inputs \cite{lillicrap2015continuous,silver2014deterministic}. Refer to Figure \ref{fig:algorithm comparison}, DDPG may be constructed by directly adding target networks and experience replay buffer to the vanilla dHDP.

The DDPG is a policy gradient method. It uses mini-batch data $(x_k,u_k,x_{k+1})$ sampled from the experience replay buffer. Its critic ${Q}_i$ in the {$i$}th iteration is trained via gradient descent to reduce the Bellman error represented as an $\ell^2$ loss,
\begin{equation}
L_{DDPG}=\frac{1}{N} \sum_{k=1}^N\left(y-{Q}\left(x_k, u_k\right)\right)^2,
\end{equation}
where the target value
\begin{equation}
 y=R(x_k,u_k)+ \gamma {Q}^{\prime}\left(x_{k+1}, \pi^{\prime}\left(x_{k+1}\right)\right),
 \label{equ:target ys}
\end{equation}
and $N$ is the mini batch size, $k$ is the sample index from the experience buffer, $Q'$ and $\pi'$ are target critic network and target actor network, respectively. 
The regression target $y$ in (\ref{equ:target ys}) is to improve learning stability. Two different mechanisms, namely the ``hard" (or ``periodic") and ``soft", are used in updating the target network (output $y$) where the  ``hard" directly copies the weights from the respective actor or critic networks \cite{lillicrap2015continuous}, and the ``soft"  only copies a small portion, for example, 5\% of the respective  weights from the actor or critic network, respectively, at each  iteration \cite{lillicrap2015continuous,fujimoto2018addressing}.

The actor network is updated based on the following policy gradient estimator,
\begin{equation}
g_{_{DDPG}}=\frac{1}{N}\sum_{k=1}^N \nabla_\pi {Q}\left(x_k, \pi(x_k)\right).
\end{equation}

\section{Phased Actor in Actor-Critic (PAAC) Reinforcement Learning}
\label{sec:PAAC}

In this section, we introduce the new PAAC  that can be integrated into the actor network of an actor-critic reinforcement learning structure. By doing so, we expect enhanced performance of the respective actor-critic methods. 

\subsection{Background}
\label{sec:background}

We study reinforcement learning control problems in which an agent 
chooses actions by interacting with the environment over a sequence of time steps in order to minimize a cumulative cost.
Consider a general nonlinear discrete-time system
\begin{equation}
    {x}_{k+1}=f({x}_k,{v}_k), k=0,1,...
\label{eq:system environment}
\end{equation}
where $k$ is the time index of system dynamics, $x_k \in \mathbb{R}^n$ is the state from the environment $E$ and $u_k \in \mathbb{R}^m$ is the action applied to the influence the environment. The system dynamic transition model  $f$ is unknown. 

For system (\ref{eq:system environment}) and a positive definite stage cost $R\left(x_{k+j}, u_{k+j}\right)$, we consider the infinite horizon discounted cost-to-go ${Q}(x_k,u_k)$ for policy $\pi(\cdot)\in\mathbb{R}^n \times \mathbb{R}^m$, $u_k = \pi(x_k)$, applied from ($k+1$)th step onward
\begin{equation}
{Q}\left(x_k, u_k\right)=\mathbb{E}_{R,x\sim E,u\sim \pi}[\sum_{j=0}^{\infty} \gamma^j R\left(x_{k+j}, u_{k+j}\right)].
\label{equ: true value of Q}
\end{equation}
For deterministic policies which are usually the case for continuous control applications, 
\begin{equation}
{Q}\left(x_k, u_k\right)=R\left(x_k, u_k\right)+\mathbb{E}_{R,x\sim E}[\sum_{j=1}^{\infty} \gamma^j R\left(x_{k+j}, \pi(x_{k+1})\right)]. 
\label{equ:HJB equation with expectation}
\end{equation}
We thus have the following recursive relationship for the cost-to-go function under policy $\pi$.
\begin{equation}
{Q}\left(x_k, u_k\right)=R\left(x_k, u_k\right)+\gamma {Q}\left(x_{k+1},\pi(x_{k+1})\right).
\label{equ:HJB equation}
\end{equation}

The objective of optimal control is to find a control policy that can stabilize system (\ref{eq:system environment}) and minimize the cost-to-go in (\ref{equ:optimal control h}), or (\ref{equ:dt hjb}) for deterministic policies. 

According to the Bellman optimality principle, the optimal cost-to-go satisfies the following relationship,
\begin{equation}
    {Q}^*(x_k,u_k) =R(x_k,u_k)+\gamma {Q}^*(x_{k+1},\pi^*(x_{k+1})),
\label{equ:one step Q}
\end{equation}
and the optimal control law $\pi^*$ can be expressed as 
\begin{equation}
   \pi^*(x_k) = arg \min _{u_k}{Q}^*(x_{k},u_{k}),
    \label{equ:optimal control h}
\end{equation}
Combining (\ref{equ:one step Q}) and (\ref{equ:optimal control h}),
\begin{equation}
    {Q}^*(x_k,u_k) =R(x_k,u_k)+\gamma \min _{u_{k+1}} {Q}^*(x_{k+1},u_{k+1}),
    \label{equ:dt hjb}
\end{equation}
where ${Q}^*\left(x_k, u_k\right)$ is the state-action value function corresponding to the optimal control policy $\pi^*\left(x_k\right)$. This equation reduces to the Riccati equation as a special case in an LQR setting, which can be efficiently solved.

We need the following definition and assumption to develop our results.

\textbf{Definition 1.} (Stabilizable System) A nonlinear dynamical system is said to be stabilizable on a compact set $\Omega \in \mathbb{R}^n$, if for all initial states $x_0 \in \Omega$, there exists a control sequence $u_0, u_1, \ldots, u_k, \ldots$
such that the state $x_k \rightarrow 0$ as $k \rightarrow \infty$.

\begin{assumption}
\label{as:assumption 1}
System (\ref{eq:system environment}) is controllable and stabilizable. The system state $x_k=0$ is an equilibrium  of the system  under the control $u_k=\pi\left(x_k\right)=0$ for $x_k= 0$, i.e., $f(0,0)=0$. The feedback control sequence $u_k$ is determined from control policy $\pi$ represented by the actor neural network, and in the most general case is bounded by actuator saturation. 
\end{assumption}

\begin{assumption}
\label{as:assumption 2}
The stage cost function $R\left(x_k, u_k\right)$ is finite, continuous in $x_k$ and $u_k$, and positive semi-definite with $R(x_k,u_k)=0$ if and only if $x_k=0$ and $u_k$=0..
\end{assumption}

\subsection{Phased Actor}

This study introduces a phased actor in actor-critic (PAAC) RL. As such, PAAC performs policy evaluation and policy improvement alternately just as typical actor-critic methods do. But, PAAC entails a new mechanism in updating the actor network. To put the discussion in context, in typical actor-critic learning \cite{sutton2018reinforcement},
a TD error is used to guide learning in the critic network, as well as the actor network, especially stochastic polices. 
In other actor-critic  algorithms such dHDP, DDPG, and TD3, the policy gradient is usually based on the $Q$ value directly.
Refer to Figure 1, PAAC is a new actor learning mechanism for deterministic policy gradient. It uses a combination of the $Q$ value and the TD error to compute the policy gradient in actor weight updates. The $Q$ value plays a dominant role 
during early stage of learning, but TD error becomes dominant 
at later stage.  By using a transition function  as a design hyperparameter, we can control the transition from  using $Q$ value to the TD error.

In PAAC, for a deterministic control policy $ u_k = \pi(x_k)$, 
the state-action value function, or the cost-to-go function, $Q(x_k,u_k)$ for policy $\pi$ is given as 
\begin{equation}
Q\left(x_k, u_k\right)=\mathbb{E}_{R,x\sim E}[\sum_{j=0}^{\infty} \gamma^j R\left(x_{k+j}, u_{k+j}\right)],
\label{equ: PAAC true value of Q}
\end{equation}
which has a recursive form as 
\begin{equation}
Q\left(x_k, u_k\right)=R\left(x_k, u_k\right)+\gamma Q\left(x_{k+1},\pi(x_{k+1})\right).
\label{equ: PAAC HJB equation}
\end{equation}

Let $\theta$ be the set of weight parameters of the actor network, namely the PAAC policy $\pi$ can be denoted as $\pi(x_k|\theta)$, and $B$ be the memory buffer which stores finite number of transitions $(x_k,u_k,R(x_k,u_k),x_{k+1})$ with the oldest samples being removed first as new samples coming in. We define the policy gradient  as
\begin{equation}
g= \mathbb{E}_{R,x_k \sim B} [\nabla_\theta \Psi^k(\theta)],
\label{equ:actor loss}
\end{equation}
where 
\begin{equation}
\Psi^k(\theta) =
\begin{cases}Q\left(x_{k}, \pi\left(x_{k}|\theta\right)\right) &\text { if } \omega \leq M(k) \\ 
\\
\delta(\theta)
&\text { if } \omega>M(k)
\end{cases}.
\label{equ:actor loss cases}
\end{equation}
The TD error $\delta(\theta)$ in the above is a function of $(x_k, u_k, R(x_k,u_k), x_{k+1},\pi(x_k|\theta))$, and
\begin{equation}
\delta(\theta) = Q\left(x_{k}, \pi\left(x_{k}|\theta\right)\right) - y,   
\end{equation}
where the target 
\begin{equation}
y=\gamma Q^{\prime}(x_{k+1}, \pi^{\prime}\left(x_{k+1}\right))+R\left(x_k, u_k\right), 
\end{equation}
is the same as that used in training the critic network. $M(k)$ is designed as a monotonically decreasing function from 1 to 0 as learning proceeds, and  $\omega \in [0,1]$  is a uniformly distributed random variable. During each update, the probability of choosing $Q$ value and TD error is $M(k)$ and  $1- M(k)$, respectively. Therefore, the gradient $g$ gradually changes from favoring $Q$ value that encourages exploration to favoring TD error that helps reduce variance as shown in section \ref{sec:variance analysis}.
With the policy gradient determined from  Equation (\ref{equ:actor loss}), the weight $\theta$ of the actor network can be updated using standard procedures such as stochastic gradient descent (SGD) or ADAM optimizer. The implementation of PAAC is described in Algorithm \ref{alorithm box}.

\begin{remark} 
As PAAC is developed within a deterministic policy gradient setting, the $Q$ valued-based policy gradient estimation in DPG algorithms, such as DDPG, TD3 and dHDP, can be directly replaced by PAAC in the actor policy updates.
Note that, PAAC is in no conflict with other policy gradient actor-critic learning, no matter their respective policies are determined directly from $Q$ value or from TD error. 
In essence, 
PAAC only replaces them with a composite value that transitions from $Q$ to TD error as learning proceeds.
\end{remark}

\begin{algorithm}
\caption{The PAAC Algorithm }
\label{alorithm box}
\begin{algorithmic}
\State Initialize critic $Q(\cdot)=0$, actor $\pi(\cdot)$ by pytorch.nn.init
\State Initialize target network $Q^\prime$ and $\pi^\prime$ with weights copied from $Q$ and $\pi$
\State Initialize replay buffer $B$
\State Set the transition function $M(k)$
\For{episode = 1, $E_{\#}$}       (with episode length $T$)
    \State Receive initial observation state $x_0$
    \For{s = 1, T}
        \State Select action $u_s=\pi(x_s) + \mathcal{N}_s$ with random exploration noise $\mathcal{N}_s$
        \State Execute action $u_s$ and observe cost $R(x_s,u_s)$ and  new state $x_{s+1}$
        \State Store transition $(x_s,u_s,R(x_s,u_s),x_{s+1})$ in $B$
        \State Sample a random minibatch of $N$ transitions $(x_k,u_k,R(x_k,u_k),x_{k+1})$ from $B$
            \State Set $y=\gamma Q^{\prime}(x_{k+1}, \pi^{\prime}\left(x_{k+1}\right))+R\left(x_k, u_k\right)$
        \State Update critic by minimizing the loss: $L = \frac{1}{N}\sum (Q\left(x_k,\pi(x_k)\right)-y)^2$
        \State Generate $\omega$ from a uniform distribution
        \If{$\omega<M(k)$}
            \State Update the actor policy using $Q$ in (\ref{equ:actor loss cases})
        \Else
            \State Update the actor policy using $\delta(\theta)$ in (\ref{equ:actor loss cases})
        \EndIf
    \EndFor
\EndFor
\end{algorithmic}

\end{algorithm}

We are now in a position to introduce some inherent properties associated with PAAC. They are important for the development of further analytical results on PAAC learning performance. Next, we first show  that the policy gradient used in PAAC minimizes the same $Q$ value even though PAAC also depends on TD error. Then we show that PAAC helps reduce the variance of policy gradient.  
\begin{theorem}
\label{theorem: PAAC minimize Q}
Consider the state-action value 
$Q(x_k,u_k)$ as in (\ref{equ: PAAC HJB equation}), and the control policy $\pi(x_k|\theta)$, the parameters of which are updated based on the policy gradient estimator in (\ref{equ:actor loss}). 
Then we have the following
\begin{equation}
\label{equ:weight update paac}
    \mathbb{E}_{R,x_k \sim B} [\nabla_\theta \Psi^k(\theta)] =  \mathbb{E}_{R,x_k \sim B} [\nabla_\theta Q\left(x_{k}, \pi\left(x_{k}|\theta\right)\right)].
\end{equation}
\end{theorem}

\noindent{\textit{Proof}}. 
First, we investigate the policy gradient using PAAC. Based on (\ref{equ:actor loss cases}),
\begin{equation}
    \begin{aligned}
       \mathbb{E}_{R,x_k \sim B}&[\nabla_\theta \Psi^k(\theta)]\\
       &=M(k)\mathbb{E}_{R,x_k \sim B}[\nabla_\theta Q\left(x_{k}, \pi\left(x_{k}|\theta\right)\right)]\\
       &+(1-M(k))\mathbb{E}_{R,x_k \sim B}[\nabla_\theta \delta(\theta)].\\
    \end{aligned}
\label{equ:gradient of PAAC}
\end{equation}  
Then, we examine the gradient of the TD error over policy parameters. Since target value $y$ does not depend on the current policy $\pi$, we have $\mathbb{E}_{R,x_k \sim B}\left[\nabla_\theta y \right]=0$. Therefore,
\begin{equation}
\label{equ:PAAC gradient of td}
\begin{aligned}
&\mathbb{E}_{R,x_k \sim B} \left[\nabla_\theta \delta(\theta)\right]\\
& = \mathbb{E}_{R,x_k \sim B} \left[\nabla_\theta \left(  Q(x_{k}, \pi(x_{k}|\theta)) -y\right)\right] \\
& =\mathbb{E}_{R,x_k \sim B} \left[\nabla_\theta   Q(x_{k}, \pi(x_{k}|\theta))\right] +\mathbb{E}_{R,x_k \sim B} \left[\nabla_\theta y \right] \\
& =\mathbb{E}_{R,x_k \sim B} \left[\nabla_\theta  Q(x_{k}, \pi(x_{k}|\theta))\right].\\
\end{aligned}
\end{equation}
Combining (\ref{equ:gradient of PAAC}) and (\ref{equ:PAAC gradient of td}) leads to policy gradient of PAAC,
\begin{equation}
    \begin{aligned}
       \mathbb{E}_{R,x_k \sim B}&[\nabla_\theta \Psi^k(\theta)]\\
       &= M(k)\mathbb{E}_{R,x_k \sim B}[\nabla_\theta Q\left(x_{k}, \pi\left(x_{k}|\theta\right)\right)]\\
       &+\big(1-M(k)\big)\big(\mathbb{E}_{R,x_k \sim B}[\nabla_\theta Q\left(x_{k}, \pi\left(x_{k}|\theta\right)\right)]\big)\\
       &=\mathbb{E}_{R,x_k \sim B}[\nabla_\theta Q\left(x_{k}, \pi\left(x_{k}|\theta\right)\right)].
    \end{aligned}
\end{equation}
The proof is complete.

\begin{remark} 
According to Theorem \ref{theorem: PAAC minimize Q}, the policy parameter updates based on PAAC make adjustments in a direction to minimize the $Q$ value. We thus can use the following iterative procedure that involves policy improvement and policy update to solve the Bellman optimality equation in (\ref{equ:dt hjb}). Even though this iterative procedure is the same as typical actor-critic learning, the actor parameter update differs from other methods as PAAC is used in learning the actor network. 

According to Algorithm \ref{alorithm box}, for $i=0,1,2, \ldots$, PAAC proceeds with the following policy update and policy evaluation steps iteratively.
\begin{equation}
Q_{i+1}\left(x_k, u_k\right)=R\left(x_k, u_k\right)+\gamma Q_i\left(x_{k+1}, \pi_i\left(x_{k+1}\right)\right),
\label{equ:PAAC approximation error statisfication}
\end{equation}
and
\begin{equation}
\pi_{i}\left(x_k\right)=\arg \min _{u_k} Q_i\left(x_k, u_k\right).
\label{eq:PAAC policy update}
\end{equation}
Or by combining (\ref{equ:PAAC approximation error statisfication}) and (\ref{eq:PAAC policy update}), we have
\begin{equation}
Q_{i+1}\left(x_k, u_k\right)=R\left(x_k, u_k\right)+\gamma \min _{u_{k+1}} Q_i\left(x_{k+1}, u_{k+1}\right).
\label{equ: PAAC iteration equ}
\end{equation}

\end{remark}

\subsection{PAAC Reduces Variance of Policy Gradient }
\label{sec:variance analysis}
We now analyze the variance of the policy gradient by PAAC. The following analysis shows that PAAC does not make the variance of the policy gradient worse during early learning phase, but it reduces the variance toward late stage of learning.

\begin{theorem} 
Let $y_i=\gamma Q_i^{\prime}(x_{k+1}, \pi_i^{\prime}\left(x_{k+1}\right))+R\left(x_k, u_k\right)$,  and $\delta_i(\theta) = Q_i\left(x_{k}, \pi_i\left(x_{k}|\theta\right)\right) – y_i $. Assume $\{x_k\}, k ={1, ..., N}$, is randomly drawn from the memory buffer $B$. Additionally, assume that $Q_0 = var(Q_0)=0$. We then have that
$var(\nabla_\theta  Q_i(x_{k})) \geq var(\nabla_\theta  \delta_i(\theta))$.
\end{theorem}

\noindent{\textit{Proof}}. 
Let $Q_i(x_{k}, \theta)$ be shorthanded for $Q_i(x_{k}, \pi_i(x_{k}|\theta))$ in (\ref{equ:PAAC approximation error statisfication}). Inspect the variance of the policy gradient  based on $Q$ value and TD error, respectively.
\begin{equation}
\begin{aligned}
&var(\nabla_\theta  Q_i(x_{k},\theta))\\
&= \mathbb{E}[(\nabla_\theta Q_i(x_{k},\theta))^2] - (\mathbb{E}[\nabla_\theta Q_i(x_{k},\theta)])^2.\\
&var(\nabla_\theta  \delta_i(\theta))\\
&= \mathbb{E}[(\nabla_\theta ( Q_i(x_{k},\theta)-y_i))^2] - (\mathbb{E}[\nabla_\theta ( Q_i(x_{k},\theta)-y_i)])^2.  
\end{aligned}
\end{equation}
According to (\ref{equ:PAAC gradient of td}),
\begin{equation}
(\mathbb{E}[\nabla_\theta( Q_i(x_{k},\theta)-y_i)])^2=(\mathbb{E}[\nabla_\theta  Q_i(x_{k},\theta)])^2.
\label{equ:variance simplified}
\end{equation}
By using (\ref{equ:variance simplified}) we have 
\begin{equation}
\begin{aligned}
&var\big(\nabla_\theta   Q_i\left(x_{k},\theta\right)\big)  - var\big(\nabla_\theta  \delta_i\left(\theta\right)\big)\\
&=\mathbb{E}\big[\big(\nabla_\theta Q_i\left(x_{k},\theta \right)\big)^2\big]-\big(\mathbb{E}\big[\nabla_\theta Q_i(x_{k},\theta)\big]\big)^2\\
&-\Big(\mathbb{E}\big[\big(\nabla_\theta( Q_i(x_{k},\theta)-y_i)\big)^2\big]-\big(\mathbb{E}\big[\nabla_\theta\big( Q_i(x_{k},\theta)-y_i\big)\big]\big)^2\Big)\\
&= \mathbb{E}\big[\big(\nabla_\theta Q_i\left(x_{k},\theta \right)\big)^2\big]-\mathbb{E}\big[\big(\nabla_\theta( Q_i(x_{k},\theta)-y_i))\big)^2\big]\\
&-\big(\mathbb{E}\big[\nabla_\theta Q_i(x_{k},\theta)\big]\big)^2+\big(\mathbb{E}\big[\nabla_\theta Q_i(x_{k},\theta)\big]\big)^2 \\
&=\mathbb{E}\big[\big(\nabla_\theta Q_i\left(x_{k},\theta \right)\big)^2\big]-\mathbb{E}\big[\big(\nabla_\theta( Q_i(x_{k},\theta)-y_i)\big)^2\big].\\
\label{equ:variance difference between q and td}
\end{aligned}
\end{equation}
Note, the target networks are updated every $d$ steps, hence we can rewrite $y_i$ as
\begin{equation}
  y_i  = Q_{i-d}(x_{k+1}, \pi_{i-d}\left(x_{k+1}\right)) +R(x_k,u_k).
\end{equation}

From here we prove the theorem in two cases.

\noindent{\textbf{Case 1}}: For $i=1,2,3...$, and $Q_i$ updates are in the early stage of learning. We have 
\begin{equation}
    \mathbb{E}\big[\big(\nabla_\theta Q_i\left(x_{k},\theta \right)\big)^2\big]=\mathbb{E}\big[\big(\nabla_\theta( Q_i(x_{k},\theta)-y_i)\big)^2\big],
\end{equation}
as $y_i$ does not depend on $\theta$. Therefore,  $\nabla_\theta y_i = 0$ . Consequently,  
\begin{equation}
    var\big(\nabla_\theta   Q_i\left(x_{k},\theta\right)\big)  - var\big(\nabla_\theta  \delta_i\left(\theta\right)\big) = 0
\end{equation}

\noindent{\textbf{Case 2}}: For $i \rightarrow \infty$, according to Theorem \ref{theorem: convergence}, $Q_i$ approaches optimality. we thus have
\begin{equation}
\begin{aligned}
    y_i &= \gamma Q_{i-d}(x_{k+1}, \pi_{i-d}\left(x_{k+1}\right))+ R(x_k,u_k)\\
        &= \gamma Q_{\infty}(x_{k+1}, \pi_{\infty}\left(x_{k+1}\right))+ R(x_k,u_k)\\
        & = Q_{\infty}(x_{k}, \pi_{\infty}\left(x_{k}\right)).
\end{aligned}
\end{equation}
Also, since $Q_i(x_k,\theta)-y_i = 0$ as $i \rightarrow \infty$, (\ref{equ:variance difference between q and td}) becomes
\begin{equation}
    var\big(\nabla_\theta   Q_i\left(x_{k},\theta\right)\big)  - var\big(\nabla_\theta  \delta_i\left(\theta\right)\big) \geq 0
\end{equation}

\section{Analysis of Learning Performance}
\label{sec:performance guarantee}
The previous section has provided basic constructs of PAAC in terms of  how it is used in updating the actor weights, and how the policy evaluation and policy improvement take place according to  (\ref{equ: PAAC iteration equ}). We now qualitatively analyze PAAC for properties associated with the convergence, solution optimality, and 
stability.

\begin{theorem}
\label{theorem: convergence} 
Let Assumptions \ref{as:assumption 1} and \ref{as:assumption 2} hold.  Let {$Q_i$} be the sequence of estimated $Q$ values starting from $Q_0 = 0$. For policy ${\pi_i}$, its actor network weights are updated based on the policy gradient estimator (\ref{equ:actor loss}), and the controls are bounded by the output function of the action network. Then 

(1) there is an upper bound $Y$ such that $0 \leq Q_i(x_k, u_k) \leq Y$, for $i =1,2,...$. 

(2) ${Q_i}$ is a non-decreasing sequence satisfying $Q_i(x_k,u_k) \leq Q_{i+1}(x_k,u_k), \forall i$. 

(3) the limit of the sequence, $Q_{\infty}\left(x_{k}, u_{k}\right)=\lim _{i \rightarrow \infty} Q_{i}\left(x_{k}, u_{k}\right)$,  satisfies 
\begin{equation}
Q_{\infty}\left(x_{k}, u_{k}\right)=R\left(x_{k}, u_{k}\right)+\gamma \min_{u_{k+1}} Q_{\infty}\left(x_{k+1}, u_{k+1}\right).
\label{eq:theorem 2}
\end{equation}

(4) the $Q$-value sequence $Q_{i}\left(x_{k}, u_{k}\right)$ and the corresponding policy $\pi_{i}\left(x_{k}\right)$, with $\pi_{\infty}\left(x_{k}\right)=\lim _{i \rightarrow \infty} \pi_{i}\left(x_{k}\right)$, converge to the optimal value $Q^{*}$ and optimal policy $\pi^{*}$, respectively:
\begin{equation}
\pi_{\infty}\left(x_{k}\right) =\pi^{*}\left(x_{k}\right),
\label{equ:t3 1}
\end{equation}
\begin{equation}
Q_{\infty}\left(x_{k}, u_{k}\right) =Q^{*}\left(x_{k}, u_{k}\right).
\label{equ:t3 2}
\end{equation}
\end{theorem}

\noindent{\textit{Proof}}.
(1) Let $\eta(x_k)$ be a deterministic control policy represented by a neural network which is  a continuous mapping from $x_k$
in stochastic environment $E$.
Let $ Z_0(\cdot) = 0$, and $Z_i$ be updated by
\begin{equation}
Z_{i+1}\left(x_{k}, u_{k}\right)=R\left(x_{k}, u_{k}\right)+\gamma Z_{i}\left(x_{k+1}, \eta\left(x_{k+1}\right)\right), \\
\end{equation}
Thus, 
$Z_{1}\left(x_{k}, u_{k}\right)=R\left(x_{k}, u_{k}\right)$.

According to Lemma 2 in \cite{gao2023reinforcement}, we obtain

\begin{equation}
\begin{aligned}
&Z_{i+1}\left(x_{k}, u_{k}\right) \\
&= \sum_{j=0}^{i} \gamma^{j} R\left(x_{k+j},\eta\left(x_{k+j}\right)\right)
\leq \sum_{j=0}^{\infty} \gamma^{j} R\left(x_{k+j}, \eta\left(x_{k+j}\right)\right).
\end{aligned}
\end{equation}
If Assumption \ref{as:assumption 1} holds, $R(x_{k+j}, u_{k+j})$ is bounded, there exists an upper bound $Y$ such that
\begin{equation}
\sum_{j=0}^{\infty} \gamma^{j} R\left(x_{k+j}, \eta\left(x_{k+j}\right)\right)  \leq Y,
\end{equation}
According to Lemma 1 in \cite{gao2023reinforcement}, as $Q_{i+1}$ is the result of minimizing the right-hand side of (\ref{equ: PAAC iteration equ}), we have
\begin{equation}
Q_{i+1}\left(x_{k}, u_{k}\right) \leq Z_{i+1}\left(x_{k}, u_{k}\right) \leq Y, \forall i.
\label{equ:Q leq Z}
\end{equation}

(2) Define a value sequence ${\Phi_i}$ as
\begin{equation}
\Phi_{i+1}\left(x_{k}, u_{k}\right)=R\left(x_{k}, u_{k}\right)+\gamma \Phi_{i}\left(x_{k+1}, \pi_{i+1}\left(x_{k+1}\right)\right),
\label{eq:theorem 1 1}
\end{equation}
and ${\Phi_0}=Q_0 = 0$. In the following, a shorthand notation is used for  $\Phi_i(x_{k+1},\pi_{i+1})= \Phi_i(x_{k+1},\pi_{i+1}(x_{k+1})) $.

Since $\Phi_0\left(x_k, u_k\right) = 0$ and $Q_1\left(x_k, u_k\right)=R(x_k,u_k)$, and $R_k$ is positive semi-definite under Assumption \ref{as:assumption 2},  
\begin{equation}
\Phi_0\left(x_k, u_k\right) \leq Q_1\left(x_k, u_k\right).
\end{equation}
From (\ref{equ:PAAC approximation error statisfication}) and (\ref{eq:theorem 1 1}), we get
\begin{equation}
\begin{aligned}
Q_{i+1}\left(x_k, u_k\right) & -\Phi_i\left(x_k, u_k\right) \\
& =\gamma\left[Q_i\left(x_{k+1}, \pi_i\right)-\Phi_{i-1}\left(x_{k+1}, \pi_i\right)\right] \geq 0.
\end{aligned}
\end{equation}
Therefore, 
\begin{equation}
\Phi_{i}\left(x_{k}, u_{k}\right) \leq Q_{i+1}\left(x_{k}, u_{k}\right).
\end{equation}
Further by using Lemma 1 in \cite{gao2023reinforcement}
\begin{equation}
Q_{i}\left(x_{k}, u_{k}\right) \leq \Phi_{i}\left(x_{k}, u_{k}\right) \leq Q_{i+1}\left(x_{k}, u_{k}\right).
\label{eq:theorem 1 conclusion}
\end{equation}
This completes the proof of Theorem \ref{theorem: convergence} (2). 

(3) From parts (1) and (2) in the above, ${Q_i}$ is a monotonically non-decreasing sequence with an upper bound. Therefore, its limit exists. Let the limit be $\lim _{i \rightarrow \infty} Q_{i}\left(x_{k}, u_{k}\right)=Q_{\infty}\left(x_{k}, u_{k}\right)$. 

 Given $i$ and for any $u_{k+1}$, according to (\ref{equ:PAAC approximation error statisfication}), there is
 \begin{equation}
Q_{i}\left(x_{k}, u_{k}\right) \leq R\left(x_{k}, u_{k}\right)+\gamma Q_{i-1}\left(x_{k+1}, u_{k+1}\right).
\end{equation}
As $Q_i$ is monotonically non-decreasing, we have
\begin{equation}
Q_{i-1}\left(x_{k}, u_{k}\right) \leq Q_{\infty}\left(x_{k}, u_{k}\right),
\label{eq:theorem 2 inequa}
\end{equation}
the following then holds
\begin{equation}
Q_{i}\left(x_{k}, u_{k}\right) \leq R\left(x_{k}, u_{k}\right)+\gamma \min_{u_{k+1}} Q_{\infty}\left(x_{k+1}, u_{k+1}\right).
\end{equation}
As $i \rightarrow \infty$, we have
\begin{equation}
Q_{\infty}\left(x_{k}, u_{k}\right) \leq R\left(x_{k}, u_{k}\right)+\gamma \min_{u_{k+1}} Q_{\infty}\left(x_{k+1}, u_{k+1}\right).
\label{eq:theorem 2 less}
\end{equation}
On the other hand, since the cost-to-go function sequence satisfies
\begin{equation}
Q_{i+1}\left(x_{k}, u_{k}\right) = R\left(x_{k}, u_{k}\right)+\gamma \min _{u_{k+1}} Q_{i}\left(x_{k+1}, u_{k+1}\right),
\end{equation}
applying inequality (\ref{eq:theorem 2 inequa}) as $i \rightarrow \infty$,
\begin{equation}
Q_{\infty}\left(x_{k}, u_{k}\right) \geq R\left(x_{k}, u_{k}\right)+\gamma \min _{u_{k+1}} Q_{\infty}\left(x_{k+1}, u_{k+1}\right).
\label{eq:theorem 2 greater}
\end{equation}
Based on (\ref{eq:theorem 2 less}) and (\ref{eq:theorem 2 greater}),  (\ref{eq:theorem 2}) is true. This completes the proof of Theorem \ref{theorem: convergence} (3).

(4) According to Theorem \ref{theorem: convergence} (3) and by using Equations  (\ref{equ:PAAC approximation error statisfication}) and (\ref{eq:PAAC policy update}), we have
\begin{equation}
\begin{aligned}
Q_{\infty}\left(x_{k}, u_{k}\right) &=R\left(x_{k}, u_{k}\right)+\gamma \min _{u_{k+1}} Q_{\infty}\left(x_{k+1}, u_{k+1}\right) \\
&=R\left(x_{k}, u_{k}\right)+\gamma Q_{\infty}\left(x_{k+1}, \pi_{\infty}\left(x_{k+1}\right)\right),
\end{aligned}
\label{equ:t3 34}
\end{equation}
and
\begin{equation}
\pi_{\infty}\left(x_{k}\right)=\arg \min _{u_{k}} Q_{\infty}\left(x_{k}, u_{k}\right).
\label{equ:t3 35}
\end{equation}
Observing (\ref{equ:t3 34}) and (\ref{equ:t3 35}), and then (\ref{equ:one step Q}) and (\ref{equ:optimal control h}), we can find that (\ref{equ:t3 1}) and (\ref{equ:t3 2}) are true. This completes the proof of Theorem \ref{theorem: convergence} (4).

\begin{theorem}
Let  Assumptions \ref{as:assumption 1} and \ref{as:assumption 2} hold, and {$Q_i$} be the sequence of estimated $Q$ values starting from $Q_0 = 0$. For policy ${\pi_i}$, its actor network weights are updated based on the policy gradient estimator (\ref{equ:actor loss}). When $Q_i$ converge to $Q_\infty$ as $\pi_i \rightarrow \pi_\infty$, then $\pi_\infty$ is a stabilizing policy.
\end{theorem}

Proof. If Assumption \ref{as:assumption 1} holds, let $\mu(x_k)$ be a stabilizing control policy, and let its cost-to-go   $\Lambda_i$ be updated by the following equation from $\Lambda_0(\cdot) = 0$,
\begin{equation}
\Lambda_{i+1}\left(x_{k}, u_k\right)=R\left(x_{k}, u_{k}\right)+\gamma \Lambda_{i}\left(x_{k+1}, \mu\left(x_{k+1}\right)\right),
\label{eq:therem 4 1}
\end{equation}
We have 
\begin{equation}
    \Lambda_{i}\left(x_{k},u_k\right) = \sum_{j=0}^{i} \gamma^{j} R\left(x_{k+j},\mu\left(x_{k+j}\right)\right),
\end{equation}
Because $\mu(x_k)$ is a stabilizing policy, if Assumption \ref{as:assumption 1} and \ref{as:assumption 2} holds, we have $x_k \rightarrow 0$ and  $R(x_k,u_k) \rightarrow 0$ as $k \rightarrow \infty$. Therefore, $\Lambda_{i}(x_k,u_k) \rightarrow 0$ as $k \rightarrow \infty$.

Next, from Lemma 1 in \cite{gao2023reinforcement}, $\pi_i$ minimizes $Q_i$, we have
\begin{equation}
\begin{aligned}
Q_{i}\left(x_k, u_k\right) \leq  \Lambda_{i}\left(x_k, u_k\right).
\end{aligned}
\end{equation}
Since $\Lambda_{i}(x_k,u_k) \rightarrow 0$ as $k \rightarrow \infty$, we have $Q_{i}(x_k,u_k) \rightarrow 0$ as $k \rightarrow \infty$. 

From Theorem \ref{theorem: convergence} (3), we obtain $R(x_k,u_k)=0$  as $k \rightarrow \infty$.
Further, under Assumption \ref{as:assumption 2}, $R(x_k,u_k)=0$ if and only $x_k = 0$, we have $x_k \rightarrow 0$ as $k \rightarrow \infty$. This completes the proof.

\begin{figure*}
    \centering
    \includegraphics[width=450pt]{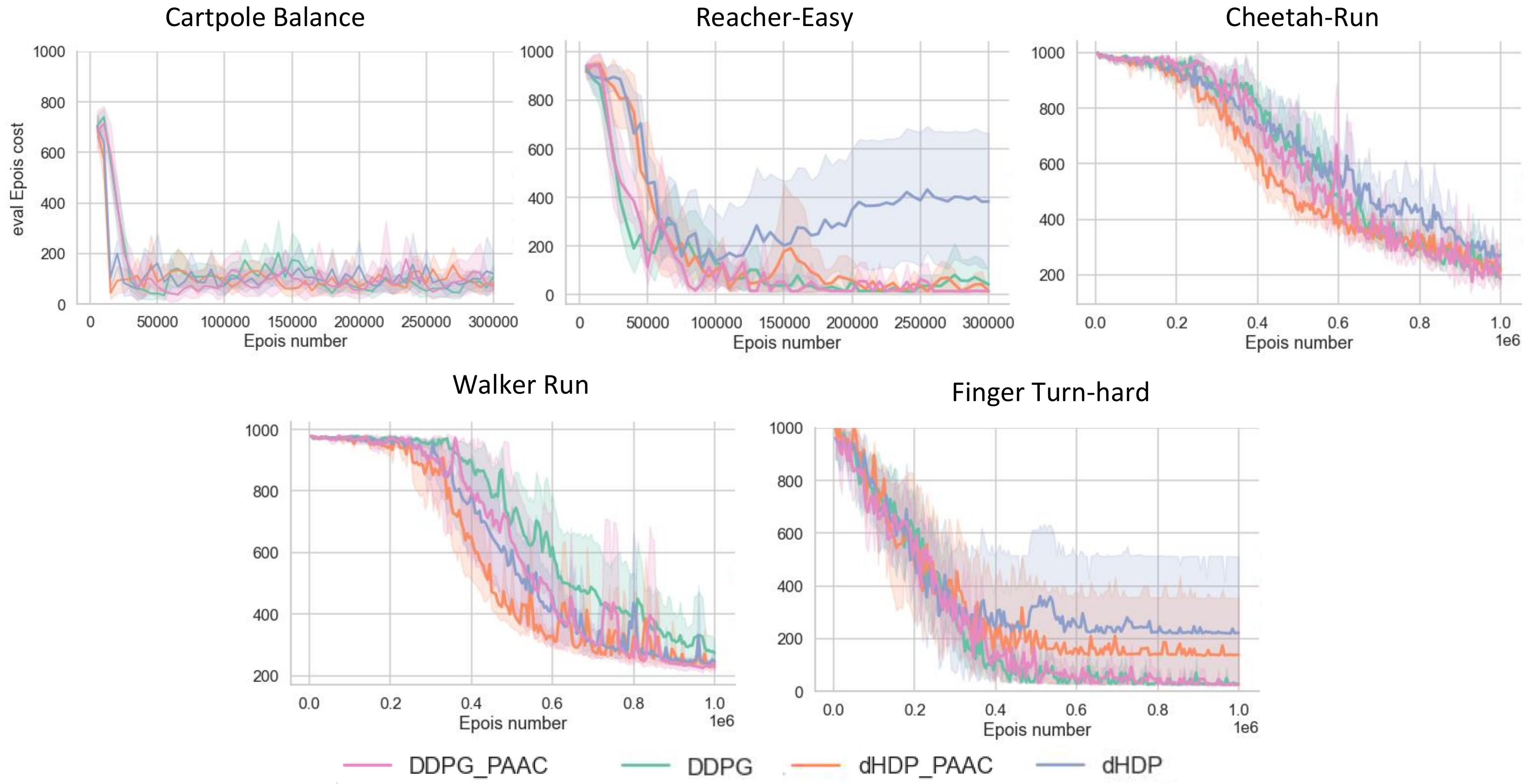}
    \caption{Learning curves of averaged total cost for benchmark study. Each learning curve is averaged over 10 different random seeds and shaded by their respective 95\% confidence interval}
    \label{fig:baseline}
\end{figure*}

\section{Results}
We provide a comprehensive evaluation of PAAC as a means of enhancing the learning performance of general policy gradient actor-critic methods. The evaluations are based on several benchmarks in DeepMind Control Suite (DMC). The evaluation tasks include the original Cartpole (Balance), Reacher (Easy),  Finger (Turn Hard), Walker (Run) and Cheetah (Run). As DMC tasks are typically set up for RL algorithms to achieve higher rewards, in the continuous state and control context, we modify the original DMC tasks from maximizing rewards to minimizing costs by using $1- R_o$ as the stage cost where $R_o$ is the original stage reward from DMC.

\subsection{General Information}
{\textbf{Network Implementations.}} 
In the implementation of the actor-critic networks, we use a two layer feed forward neural network of 256 hidden nodes for both hidden layers with rectified linear units (ReLU) between any two layers for both the actor and the critic. The final layer output units, however, use a tanh thresholding function in the actor network. The input layer of the actor has the same dimension as the observed states. The output layer of the actor has the same dimension as that of the control action. The critic uses both state and action as inputs to the first layer, and the output of the critic is the scalar $Q$ value.

{\textbf{Implementation of Target networks}}
We implement the target networks described in section \ref{sec:ddpg} (\ref{equ:target ys}) as follows.
First, the target networks have the same network architectures as the critic and actor networks, respectively. Here $Q'$ is the output of the target critic network, and $h'$ is the target action policy from the target actor network. There are two different target network updating methods implemented in this study. For the periodic updating or ``hard" 
scheme used in dHDP, the target network parameters are copied exactly from the critic and actor networks, respectively every 15 
times of parameter updates, and are held fixed before the next copy.  For ``soft updates" used in DDPG, a small portion $\tau$ of the target network parameters are copied  from the respective critic and actor networks \cite{lillicrap2015continuous,fujimoto2018addressing}, respectively, namely at each iteration, $\theta^\prime = \tau\theta+(1-\tau )\theta^\prime$ where $\theta = 0.05$ is used in the implementation.

{\textbf{Training and Evaluation.}} 
All networks are trained using the Adam optimizer with a learning rate of $10^{-3}$. 
At each time step, the network weights are updated with a mini-batch of 256 samples uniformly  drawn from the experience replay buffer containing the history of the agent-environment interaction.

Under the DMC environment, an \textbf{episode} contains $T=1000$ time steps.  Each episode of a task is initialized from a random environment condition within a small range specified in DMC. 
A \textbf{trial} corresponding to a seed is a complete training process that contains a number of consecutive episodes. For  easy tasks such as Cartpole (Balance) and Reacher (Easy), each trial is run for 0.3 million time steps, while for the difficult tasks, such as  Cheetah (Run), Walker (Run) and Finger (Turn Hard), each trial is run for 1 million time steps. All tasks are evaluated every 5000 time steps. 

All results are reported based on  10 trials (i.e., 10 seeds) where all trials are initialized with the  seeds  $(0 - 9)$. For each training trial, the first 8000 time steps are used to collect transition samples for the memory buffer. In doing so, random exploratory policies are used. Each learned policy (from 10 learning trials) is evaluated over 10 episodes where the environment is initialized with 10 seeds $(100-109)$ with no exploration noise. The total cost of the evaluation episodes are reported.

\begin{table}[ht]
\tabcolsep=4pt
\renewcommand{\arraystretch}{1.2}
\begin{tabular}{|c|c|c|c|c|c|}
\hline
    &      & DDPG   & \begin{tabular}[c]{@{}l@{}}DDPG\\ \_PAAC\end{tabular}    & dHDP     & \begin{tabular}[c]{@{}l@{}}dHDP\\ \_PAAC\end{tabular}     \\ \hline
\multirow{4}{*}{\begin{tabular}[c]{@{}l@{}} Cartpole\\ Balance\end{tabular}}                       & \begin{tabular}[c]{@{}l@{}}Total  Cost\\ $\pm$ Learning\\Variance\end{tabular}      & \begin{tabular}[c]{@{}l@{}}88.59\\ $\pm$67.78\end{tabular}   & \begin{tabular}[c]{@{}l@{}}100.00\\   $\pm$85.39\end{tabular}    & \begin{tabular}[c]{@{}l@{}}107.10\\   $\pm$115.17\end{tabular} & \begin{tabular}[c]{@{}l@{}}\textbf{82.56}\\   \textbf{$\pm$53.09}\end{tabular}   \\ \cline{2-6} 
    & Robustness & 5.64   & 3.96    & 4.58     & \textbf{2.8}      \\ \cline{2-6} 
    & AUC        & 0.130        & 0.123   & 0.123    & \textbf{0.114}    \\ \cline{2-6} 
    & Success     & 100\%        & 100\%   & 100\%    & 100\%    \\ \hline
\multirow{4}{*}{\begin{tabular}[c]{@{}l@{}}Reacher\\ Easy\end{tabular}}                           & \begin{tabular}[c]{@{}l@{}}Total  Cost\\ $\pm$ Learning\\Variance\end{tabular}  & \begin{tabular}[c]{@{}l@{}}57.33\\   $\pm$117.19\end{tabular}      & \begin{tabular}[c]{@{}l@{}}\textbf{14.81}\\  \textbf{ $\pm$4.59}\end{tabular}   & \begin{tabular}[c]{@{}l@{}}33.27\\   $\pm$46.49\end{tabular} & \begin{tabular}[c]{@{}l@{}}28.45\\   $\pm$43.76\end{tabular}   \\ \cline{2-6} 
    & Robustness & 64.89        & \textbf{7.44}    & 39.57    & 35.56    \\ \cline{2-6} 
    & AUC        & 0.151        & \textbf{0.143}   & 0.190    & 0.206    \\ \cline{2-6} 
    & Success     & 100\%        & 100\%   & 90\%     & 100\%    \\ \hline
\multirow{4}{*}{\begin{tabular}[c]{@{}l@{}}Cheetah\\ Run\end{tabular}}                            & \begin{tabular}[c]{@{}l@{}}Total  Cost\\ $\pm$ Learning\\Variance\end{tabular}  & \begin{tabular}[c]{@{}l@{}}222.81\\ $\pm$74.10\end{tabular}  & \begin{tabular}[c]{@{}l@{}}\textbf{215.59}\\   \textbf{$\pm$59.63}\end{tabular} & \begin{tabular}[c]{@{}l@{}}279.59\\   $\pm$80.14\end{tabular}  & \begin{tabular}[c]{@{}l@{}}233.40\\   $\pm$71.45\end{tabular}  \\ \cline{2-6} 
    & Robustness & 56.49        & 64.16   & 39.54    & \textbf{34.16}    \\ \cline{2-6} 
    & AUC        & 0.621        & 0.613   & 0.657    & \textbf{0.570}    \\ \cline{2-6} 
    & Success     & 100\%        & 100\%   & 100\%    & 100\%    \\ \hline
\multirow{4}{*}{\begin{tabular}[c]{@{}l@{}}Walker\\ Run\end{tabular}}                             & \begin{tabular}[c]{@{}l@{}}Total  Cost\\ $\pm$ Learning\\Variance\end{tabular}  & \begin{tabular}[c]{@{}l@{}}300.57\\ $\pm$137.05\end{tabular} & \begin{tabular}[c]{@{}l@{}}\textbf{229.37}\\  \textbf{ $\pm$18.14}\end{tabular} & \begin{tabular}[c]{@{}l@{}}254.14\\   $\pm$79.86\end{tabular}  & \begin{tabular}[c]{@{}l@{}}242.44\\   $\pm$35.30\end{tabular}  \\ \cline{2-6} 
    & Robustness & 10.33        & 10.16   & 11.34    & \textbf{9.92}     \\ \cline{2-6} 
    & AUC        & 0.690        & 0.622   & 0.606    & \textbf{0.566}    \\ \cline{2-6} 
    & Success    & 100\%        & 100\%   & 100\%    & 100\%    \\ \hline

\multicolumn{1}{|c|}{\multirow{4}{*}{\begin{tabular}[c]{@{}c@{}}Finger\\ Turn \\ hard (20)\end{tabular}}} & \begin{tabular}[c]{@{}l@{}}Total  Cost\\ $\pm$ Learning\\Variance\end{tabular}  & \begin{tabular}[c]{@{}l@{}}35.88\\   $\pm$28.26\end{tabular}        & \begin{tabular}[c]{@{}l@{}}\textbf{31.44}\\   \textbf{$\pm$24.31}\end{tabular}   & \begin{tabular}[c]{@{}l@{}}32.96\\   $\pm$28.76\end{tabular}  & \begin{tabular}[c]{@{}l@{}}33.55\\   $\pm$25.03\end{tabular} \\ \cline{2-6}

\multicolumn{1}{|c|}{}                      & Robustness & 34.96        & 27.03   & 26.60    & \textbf{23.54}   \\ \cline{2-6} 
\multicolumn{1}{|c|}{}                      & AUC        & 0.253        & 0.234    & 0.223    & \textbf{0.222}   \\ \cline{2-6} 
\multicolumn{1}{|c|}{}                      & Success     & 100\%        & 100\%   & 90\%     & 95\%     \\ \hline

\end{tabular}
\\
\caption{Performance over the last 10 evaluations over 10 trials. Comparing PAAC over different policy gradient methods.}
\label{Table:baseline}
\end{table}

\begin{figure*}
    \centering
    \includegraphics[width=450pt]{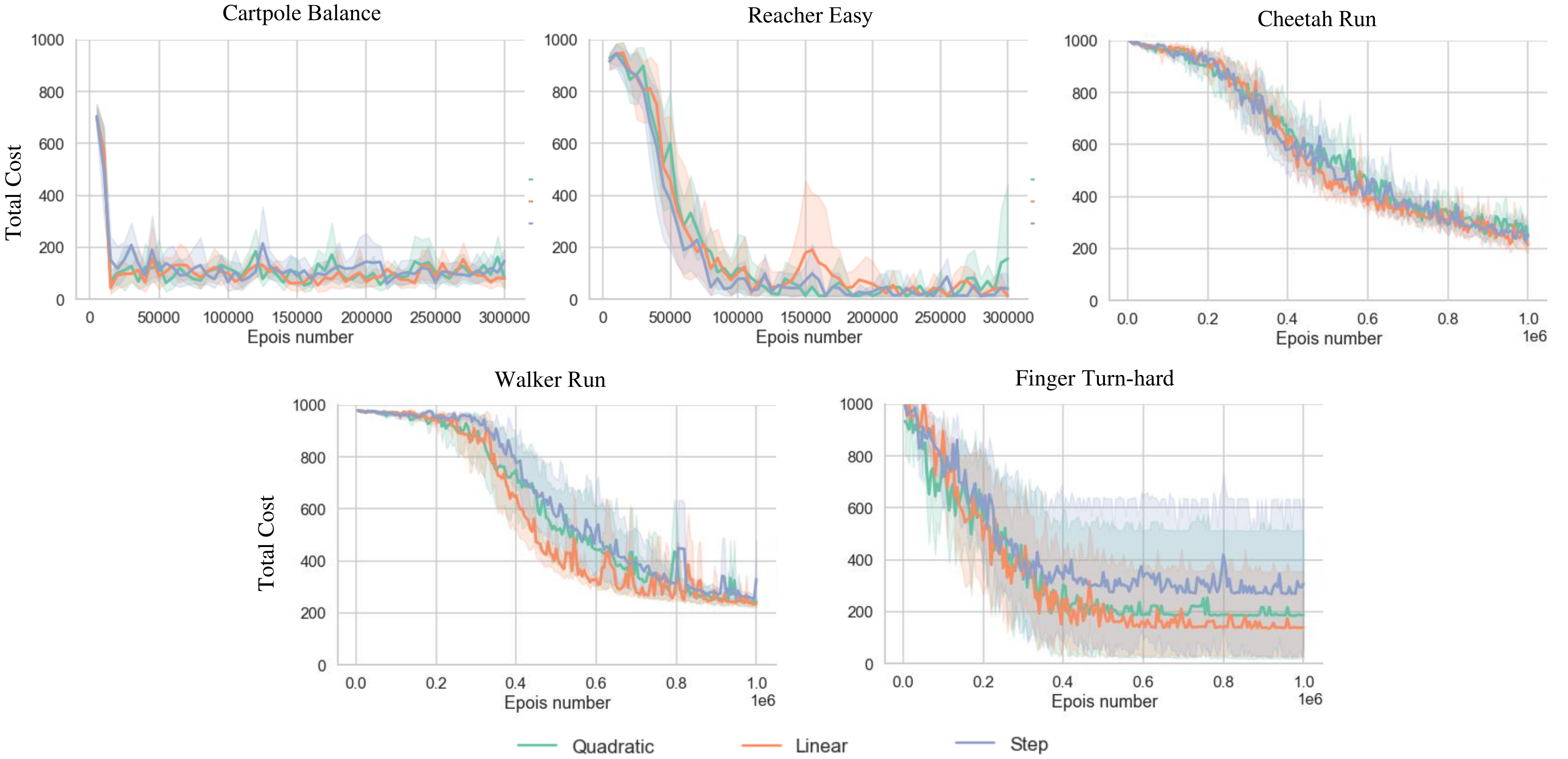}
    \caption{Learning curves of averaged total cost for different PAC switching method. Each learning curve is averaged over 10 different random seeds and shaded by their respective 95\% confidence interval}
    \label{fig:hyperparameter}
\end{figure*}

{\textbf{Benchmark Methods.}}
To demonstrate the effectiveness of PAAC, we evaluate how PAAC may improve the performance of a serious of actor-critic methods, starting from dHDP, dHDP with
experience replay and target networks, to DDPG. 
The following short-form descriptions are used  in reporting evaluation results.

1) ``Vanilla dHDP": the basic actor-critic as in \cite{si2001online}.

2) ``Updated dHDP'': also briefed as "dHDP" in result tables and figures, which is the vanilla dHDP 
with experience replay and with "hard" updates (as in \cite{si2001online}) in the target network. 

3) ``dHDP\_PAAC'': Updated dHDP with PAAC. 

4)``dHDP-ER": Updated dHDP with target networks but without experience replay. 

5)``dHDP-target": Updated dHDP with experience replay but without target networks.

6) ``DDPG'': the same as in \cite{lillicrap2015continuous}, which is equivalent to the Updated dHDP except  using  "soft"  updates in the target networks. 

7) ``DDPG\_PAAC'': namesake, DDPG \cite{lillicrap2015continuous} with PAAC.

{\textbf{Selection of Benchmark Methods.}}
We select dHDP and its variants as benchmark methods for the following considerations. 
(1) To evaluate PAAC for its efficacy it is desirable to have several benchmark methods, not just one or two or three 
(2) As shown in Related Work, we use a unified framework to show how several important actor-critic structures relate based on a basic actor-critic construct. Comparing and contrasting related methods for the evaluation of PAAC provides the opportunity to benchmark and ablate. 

Secondly, we use DDPG as another benchmark for the following reasons. 
(1) DDPG is based on DPG to also include replay buffer and deep neural networks as inspired by DQN. (2) DPG is considered a minibatch version of NFQCA \cite{silver2014deterministic} while the authors of NFQCA consider it a batch version of dHDP \cite{hafner2011reinforcement}.
Thus, we not only have a series of related methods, but also include DDPG, a popular DRL algorithm.

Lastly, we also note that a few recent actor-critic DRL algorithms have demonstrated good performance. These include TD3 \cite{fujimoto2018addressing}  and D4PG \cite{barth2018distributed}, to name some. TD3 uses clipped double Q-Learning of the critic network to mitigate the overestimation problem, and it uses delayed policy updates and target policy smoothing regularization to reduce variance \cite{fujimoto2018addressing}. D4PG introduces distributed computing, n-step TD error and prioritized experience replay \cite{schaul2015prioritized} to accelerate the learning process \cite{barth2018distributed}. These methods hold strong positions in DRL family of algorithms. Yet, since they use some unique features in their designs, and also since our focus of this study is to show PAAC as a general method and its potential to enhance actor-critic methods, we thus focus on evaluating PAAC using structures that retain the basic actor-critic constructs, that almost all actor-critic DRL algorithms rely on.

{\textbf{Phase Transition Function.}}
In the phase transition study, we compare three common functions for phase transition in PAAC. They include linear, quadratic, and hard switch functions.

{\textbf{Performance Measures.}}
The following measures are used in evaluating effectiveness of PAAC in enhancing several actor-critic methods. 

``Total Cost" : The average total cost from the last 10 evaluation episodes over 10 trials.

``Learning Variance": It measures learning performance and is computed as follows. First, the last 10 evaluations of 10 learned policies (corresponding to 10 learning trials) are considered. This results in 100 evaluations. For each of the 100 evaluations, an averaged total cost is computed based on 10 random environment seeds. 
Next, the standard deviation of those 100 averaged total cost is recorded as the Learning Variance score.

``Robustness": It measures generalization of learned policies. It is computed as follows.  First, for each of the last 10 evaluation episodes for a given learned policy, 10 environment seeds are considered. The standard deviation of the resulted 100 total costs is then obtained. Next, repeat the previous step for the remaining 9 learned policies. The Robustness score is computed from averaging all 10 standard deviations corresponding to the 10 learned policies.

``AUC": The area under a learning curve of the total cost, which is used to measure learning speed. The smaller the AUC, the faster the learning speed.

``Success" : The success rate, which is the percentage that the learning goal has been achieved in the last 10 evaluations.

\subsection{PAAC Improves Benchmark Methods}
\label{sec:baseline study}

We compare benchmark algorithms with PAAC integrated into the respective benchmark algorithms.  Figure \ref{fig:baseline} and Table \ref{Table:baseline} are summaries of the evaluation results. We also highlight (using boldface) the best performing algorithm in each task in  Table \ref{Table:baseline} where it reveals that PAAC has enabled the best performance in all comparisons. In obtaining all comparison results, PAAC uses the linear transition function.

Overall, PAAC enhances benchmark methods 
most of the time in terms of total cost, learning variance, robustness, learning speed and success rate. Figure \ref{fig:baseline} includes  learning curves of the five tasks. The respective quantitative comparisons to the benchmarks are summarized in Table \ref{Table:baseline}.  
(1) Specifically,  PAAC improves the total cost performance in 9 8 out of 10 respective comparison pairs for 5 tasks (pair 1: dHDP and dHDP\_PAAC, pair 2: DDPG and DDPG\_PAAC). For instance, PAAC improves by 30.3\% in  total cost for DDPG  except for the Cartpole (Balance) where non-PAAC performance is 12.8\% higher than with PAAC. 
(2) PAAC improves the learning variance in 9  out of 10 comparison pairs with 47.9\% 42.8\% lower standard deviation of total cost except the Cartpole (Balance) in DDPG where DDPG\_PAAC resulted in 26\% higher standard deviation. 
(3) PAAC also shows a significant 26.6\% 26.5\% performance improvement in terms of robustness in 9  out of 10 comparison pairs except the Cheetah (Run) where DDPG\_PAAC is 14.3\% worse and Figure (Turn Hard) where dHDP\_PAAC is 36.7\% worse. 
(4) For learning speed, PAAC shows an overall 11\% 6.4\% improvement in AUC except Cheetah (Run) in dHDP\_PAAC resulted in 8.4\% higher in all comparison pairs. 
(5) PAAC also  improves success rate for all including Finger (Turn Hard) for dHDP while for all other tasks, benchmarks has reached 100\% success rate.

Results in Figure \ref{fig:baseline} corroborate  those in Table \ref{Table:baseline}. Note that the fail trials are plotted in Figure \ref{fig:baseline} while they are eliminated in statistic analysis in Table \ref{Table:baseline} as outliers and  are reflected in success rate. They further reveal that PAAC has enabled more pronounced performance improvement in challenging tasks such as  Cheetah (Run), Walker (Run) and Finger (Turn Hard) than in easy tasks such as Cartpole (Balance) and Reacher (Easy). This may be due to that easy tasks are easy to learn for  the benchmark algorithms already, that is, benchmark algorithms usually reach learning  converge sufficiently quickly. As this is still quite early in the  learning stage, and PAAC is still dominated by  $Q$ values with little contributions from the TD error. Thus, for simple tasks that benchmark algorithms can effectively handle, PAAC may not be needed in those cases. However, by tailoring the switching function to individual tasks, we may still see PAAC taking effect in improving learning performance. 

Now we inspect the performances of PAAC enhanced methods on different tasks. In Figure \ref{fig:baseline}, we see that in Cartpole (Balance) , dHDP\_PAAC outperforms DDPG\_PAAC in total cost, learning variance, while DDPG\_PAAC is better in the other 4 tasks. However, for AUC and robustness which measures generalization of a learned policy, dHDP\_PAAC is better in 4 tasks except Reacher (Easy). 
For success rate, 
the improvement is observed in Reacher (Easy) and Finger (Turn Hard) of dHDP\_PAAC, while all other tasks have already reached 100\% in benchmark methods. In general, PAAC effectively improves benchmark methods using the evaluated performance measures, but the scale of improvements depends on the tasks on hand.
The overall performance of dHDP\_PAAC and DDPG\_PAAC are comparable. DDPG\_PAAC shows advantage on total cost and success rate while dHDP\_PAAC shows advantage on robustness and learning speed. In general, neither of them outperforms the other in all measurements.

\begin{table}[h]
\renewcommand{\arraystretch}{1.2}
\vspace*{5pt}
\begin{tabular}{|c|c|c|c|c|}
\hline
      &            & Linear        & Quadratic      & Hard Switch \\ \hline
\multirow{4}{*}{\begin{tabular}[c]{@{}l@{}}Cartpole\\ Balance\end{tabular}}                       & \begin{tabular}[c]{@{}l@{}}Total  Cost\\ $\pm$ Learning\\Variance\end{tabular}  & \begin{tabular}[c]{@{}l@{}}\textbf{82.56}\\ \textbf{$\pm$53.09}\end{tabular}   & \begin{tabular}[c]{@{}l@{}}114.27\\ $\pm$87.59\end{tabular}   & \begin{tabular}[c]{@{}l@{}}115.89\\ $\pm$78.11\end{tabular}                  \\ \cline{2-5} 
      & Robustness  & \textbf{2.8 }          & 4.97           & 3.33  \\ \cline{2-5} 
      & AUC        &\textbf{ 0.114 }        & 0.122          & 0.130       \\ \cline{2-5} 
      & Success    & 100\%         & 100\%          & 100\%       \\ \hline
\multirow{4}{*}{\begin{tabular}[c]{@{}l@{}}Reacher\\ Easy\end{tabular}}    & \begin{tabular}[c]{@{}l@{}}Total  Cost\\ $\pm$ Learning\\Variance\end{tabular}  & \begin{tabular}[c]{@{}l@{}}28.45\\ $\pm$43.76\end{tabular}   & \begin{tabular}[c]{@{}l@{}}87.59\\ $\pm$195.12\end{tabular}   & \begin{tabular}[c]{@{}l@{}}\textbf{26.52}\\ \textbf{$\pm$46.67}\end{tabular}  \\ \cline{2-5} 
      & {Robustness}  & 35.56         & 69.00          & \textbf{33.12}       \\ \cline{2-5} 
      & AUC        & 0.206         & 0.206          & \textbf{0.174}       \\ \cline{2-5} 
      & Success    & 100\%         & 100\%          & 100\%       \\ \hline
\multirow{4}{*}{\begin{tabular}[c]{@{}l@{}}Cheetah\\ Run\end{tabular}}     & \begin{tabular}[c]{@{}l@{}}Total  Cost\\ $\pm$ Learning\\Variance\end{tabular}  & \begin{tabular}[c]{@{}l@{}}\textbf{233.40}\\ \textbf{$\pm$71.45}\end{tabular}  & \begin{tabular}[c]{@{}l@{}}280.13\\ $\pm$72.23\end{tabular}   & \begin{tabular}[c]{@{}l@{}}256.51\\ $\pm$70.82\end{tabular}  \\ \cline{2-5}
      & {Robustness}  & \textbf{34.16}         & 50.77          & 46.78       \\ \cline{2-5} 
      & AUC        & \textbf{0.570}         & 0.594          & 0.578       \\ \cline{2-5} 
      & Success    & 100\%         & 100\%          & 100\%       \\ \hline
\multirow{4}{*}{\begin{tabular}[c]{@{}l@{}}Walker\\ Run\end{tabular}}      & \begin{tabular}[c]{@{}l@{}}Total  Cost\\ $\pm$ Learning\\Variance\end{tabular}  & \begin{tabular}[c]{@{}l@{}}\textbf{242.44}\\ \textbf{$\pm$35.3}\end{tabular}   & \begin{tabular}[c]{@{}l@{}}244.71\\ $\pm$31.62\end{tabular}   & \begin{tabular}[c]{@{}l@{}}269.82\\ $\pm$82.86\end{tabular}  \\ \cline{2-5}
      & {Robustness}  & \textbf{9.92 }         & 10.84          & 13.9  \\ \cline{2-5} 
      & AUC        & \textbf{0.566}         & 0.602          & 0.632       \\ \cline{2-5} 
      & Success    & 100\%         & 100\%          & 100\%       \\ \hline
\multicolumn{1}{|c|}{\multirow{4}{*}{\begin{tabular}[c]{@{}c@{}}Finger\\ Turn hard\end{tabular}}} & \begin{tabular}[c]{@{}l@{}}Total  Cost\\ $\pm$ Learning\\Variance\end{tabular}  & \begin{tabular}[c]{@{}l@{}}\textbf{142.45}\\ \textbf{$\pm$305.95}\end{tabular} & \begin{tabular}[c]{@{}l@{}}190.94\\ $\pm$ 365.80\end{tabular} & \begin{tabular}[c]{@{}l@{}}258.02\\ $\pm$403.09\end{tabular} \\ \cline{2-5} 
\multicolumn{1}{|c|}{}    & {Robustness}  & 18.68         & \textbf{18.46}          & 32.38       \\ \cline{2-5} 
\multicolumn{1}{|c|}{}    & AUC        & \textbf{0.321}         & 0.345          & 0.403       \\ \cline{2-5} 
\multicolumn{1}{|c|}{}    & Success    & 90\%          & 90\%           & 90\%  \\ \hline
\end{tabular}
\\
\caption{Performance over the last 10 evaluations over 10 trials. Comparing different phased transit functions.}
\label{Table:hyperparameter}
\end{table}

\begin{figure*}
    \centering
    \includegraphics[width=450pt]{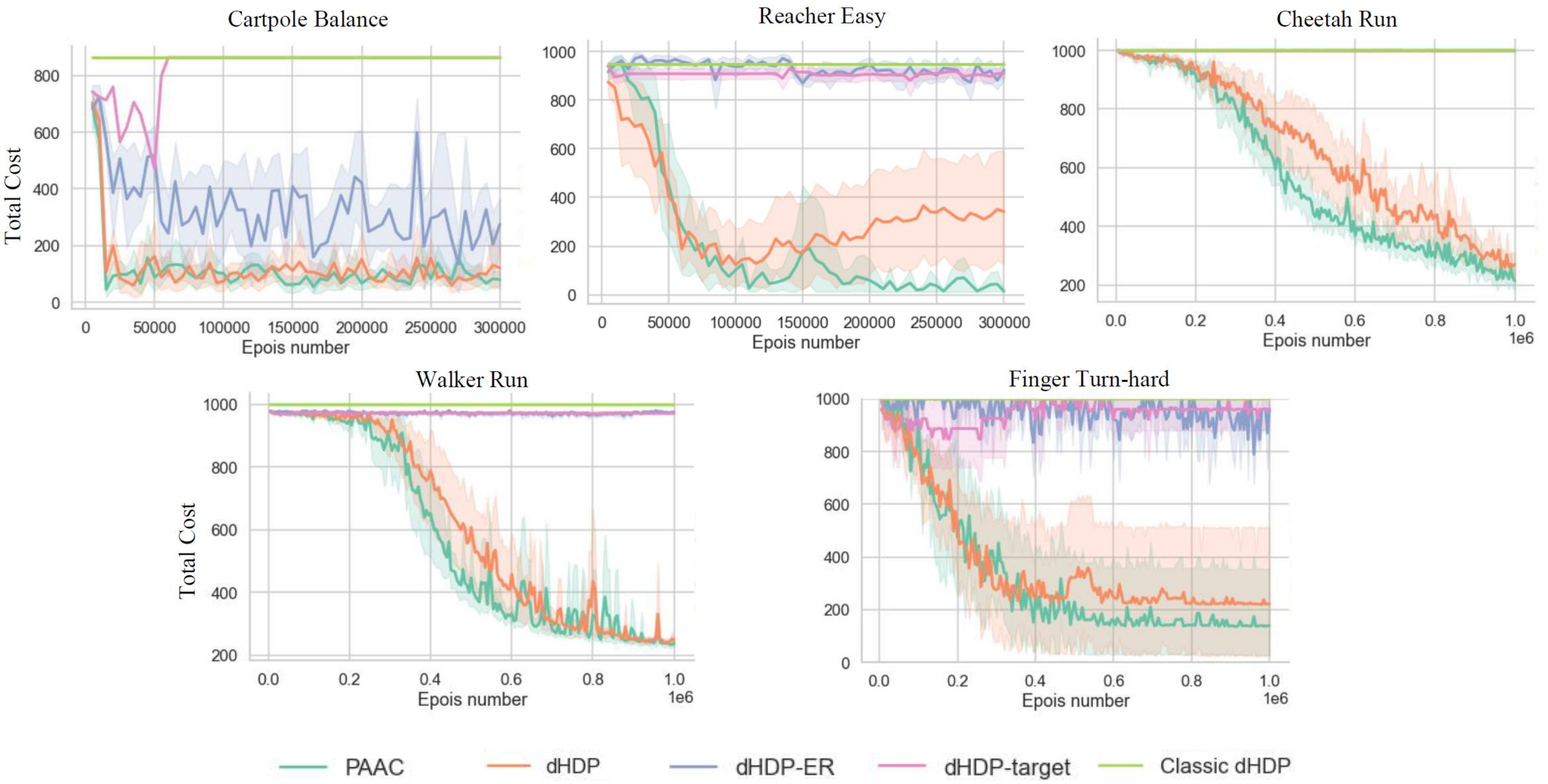}
    \caption{Learning curves of averaged total cost for ablation study. Each learning curve is averaged over 10 different random seeds and shaded by their respective 95\% confidence interval}
    \label{fig:ablation cost}
\end{figure*}

\subsection{Effect of Different Transition Functions in PAAC}
Here we compare the effect of using different phase transition functions in PAAC. Three  transitions  are evaluated including  quadratic, linear and step functions. In (\ref{equ:actor loss cases}), for the quadratic transition function, $M(k) = (1-\frac{k}{K_{total}})^2$, where $K_{total}$ is the total number of steps. For linear transition function,  $M(k) = 1-\frac{k}{K_{total}}$ is applied. For the hard switch, a step function from 1 is toggled to 0  half way through the total learning steps. 

The linear transition function is the default method for PAAC evaluations throughout this paper based on the following analysis.
Overall, results from using linear transition function  outperform those from the other two in 4 out of the 5 tasks except the Reacher (Easy) where step function tops the performance chart. More specifically, (1) linear transition beats the second best  in total cost by 38\%, 10\% and 34\% in Cartpole (Balance), Cheetah (Run) and Finger (Turn Hard), respectively. While quadratic transition almost ties (within  1\% difference) with linear transition in Walker (Run), and step function tops by 7\% in Reacher (Easy). (2) For the learning variance measure, the linear function beats the second best by 47.1\%, 6.6\% and 19.5\% in Cartpole (Balance), Reacher (Easy) and Finger (Turn Hard), respectively. While all transition functions ties  (within 1\% difference) in Cheetah (Run), quadratic function wins in Walker (Run) by 10.5\%. (3) For the  robustness measure, linear wins in Cartpole (Balance), Cheetah (Run) and Walker (Run) while quadratic function wins in Finger (Turn Hard), and step function wins in Reacher (Easy). (4) For learning speed, linear function tops in Cartpole (Balance), Walker (Run) and Finger (Turn Hard) with about 7\% lower AUC to the second best. Step function tops in Reacher (Easy) with 14\% less AUC, and almost ties with linear function (1.4\%) in Cheetah (Run). (5) Three functions achieve the same success rate in all tasks.

These evaluations point to  that the effectiveness of different transition functions varies in different environments. 
They also indicate that when to switch and how much to switch from $Q$ value to TD error depends on the progression of the learning process. If PAAC makes a  transition by attending to TD error too early, the critic may still be in the exploration stage of interacting with the environment. As such, $Q$ value based policy gradient is likely to encourage exploration.
This is what we believe has happened in the Walker (Run), which is a more challenging task. The quadratic function has worse learning speed (higher AUC) than linear function as quadratic function change to TD error quicker than linear.

On the contrary, if the actor attends to the TD error too slowly, the effect of PAAC may diminish as its depending solely on $Q$ value in policy gradient estimation may result in
In simple tasks such as Cartpole (Balance) and Reacher (Easy), different transition functions have  little impact on the final result because they all learn  very quickly during the early learning stages.

\subsection{Enhanced dHDP}
Figure \ref{fig:ablation cost} summarizes evaluations of different variants of the vanilla dHDP in an ablation study.  By adding two recently established performance enhancing techniques, namely experience replay and target networks, as well as the proposed PAAC, we evaluate the performance of a new, dHDP\_PAAC
 which may be viewd as an enhanced dHDP. 

Below  we evaluate the influence of each of the add-on components in enhanced dHDP. First, the vanilla dHDP struggles to learn  the  DMC tasks of this study. Note however that  the Cartploe (Balance) in DMC is a more complex task than that in   \cite{si2001online,barto1983neuronlike}. Specifically, the DMC environment allows the pole freely move about including hanging down below, while all early works  balances the cart-pole within 12 degrees from the center up position. 

With recently established performance enhancing elements (experience replay and target networks) added into vanilla dHDP, Figure \ref{fig:ablation cost} shows competitive performances of different dHDP variants. First, using target networks significantly improve learning performance in simple tasks such as  Cartpole (Balance) and Reacher (Easy). Simply adding target networks, the dHDP-ER is able to learn the Cartpole (Balance) task with 80\% success rate. The total cost is reduced to $\frac{1}{3}$  of the max total cost. In Reacher (Easy) and Finger (Turn Hard), it learns, but slowly, in the given 1 million time steps. 
However, using the target networks alone was not sufficient to enable learning of the rest more complicated tasks. 

Then we evaluate the effectiveness of dHDP-target. It shows a slightly better result than using dHDP-ER in Reacher (Easy) and Walker (Run) with about 1\% lower in both total cost and AUC. However, it is significantly outperformed by dHDP-ER in Cartpole (Balance) where dHDP-ER has 80\% success rate while dHDP-target fails on all trials. For the other tasks, dHDP-target is unable to learn. Therefore, experience replay alone still is not enough to learn all the 5 DMC tasks evaluated in this study.

Although using a single individual component still lacks satisfactory performance, using a  combination of the two components (experience replay and target networks) results in great performance boost. In Figure \ref{fig:ablation cost}, the updated dHDP shows a significant jump in performance in all performance measures in all five DMC tasks. It reaches 100\% success rate in four tasks and has 80\% success rate in Finger (Turn Hard).

Lastly, as we have discussed in section \ref{sec:baseline study}, if we inspect dHDP and its variants, PAAC even further improves the performance of updated dHDP with the best results in all tasks. 
More specifically, dHDP\_PAAC outperforms updated dHDP by 34.6\%, 46.5\%, 24.2\% and 17.3\% (14.6\%,19.6\%,7.7\%,9.1\%) in total cost, learning variance, robustness and AUC, respectively. For learning success rate in Finger (Turn Hard), PAAC also enables an improvement from 80\% to 90\%. Therefore, we have obtained an enhanced dHDP with using experience replay, target networks, and PAAC.

By comparing the enhanced dHDP with popular DDPG methods, it shows that the enhanced dHDP outperforms DDPG in 4 out of 5 tasks in all performance measurements except learning speed in Reacher (Easy) and total cost in Cheetah (Run). DDPG wins in Finger (Turn Hard) because the success rate for dHDP\_PAAC is 90\%.

\section{Conclusions}
In this work, we introduce a novel PAAC mechanism that uses both $Q$ value and one step TD error in the estimation of  policy gradient.  We provide a qualitative analysis to show the convergence of the $Q$ value in actor-critic learning and 
that PAAC helps reduce the variance in policy gradient estimates. By integrating  PAAC into two baseline actor-critic methods, namely dHDP and DDPG, we show that PAAC can significantly improve learning performance of the baseline methods.  
Even though PAAC is introduced, investigated, and evaluated on two baseline methods (dHDP  and DDPG) and their variants, it can also be implemented and integrated in other actor-critic algorithms. 
Lastly, PAAC has enabled a much enhanced dHDP with significant performance improvement over that of the original dHDP from over twenty years ago.

\ifCLASSOPTIONcaptionsoff
  \newpage
\fi



%

\bibliographystyle{IEEEtran}

\bibliography{enhanced.bib}

%





\begin{IEEEbiographynophoto}{}

\end{IEEEbiographynophoto}

\end{document}